\documentclass[letterpaper, 10 pt, conference]{ieeeconf}

\IEEEoverridecommandlockouts

\usepackage{etextools}
\usepackage{amssymb}
\usepackage{float}
\usepackage{makeidx}
\usepackage{amsmath}
\usepackage{bbm}
\usepackage{epsfig}
\usepackage{epsf}
\usepackage{psfrag}
\usepackage{verbatim}
\usepackage{color}
\usepackage{multirow}
\usepackage{tabularx}
\usepackage{booktabs}
\usepackage[tight,footnotesize]{subfigure}
\usepackage{array}
\usepackage{soul}
\usepackage{footnote}
\usepackage{cite}
\usepackage{dblfloatfix}

\usepackage{graphicx}
\graphicspath{{Figures}}
\DeclareGraphicsExtensions{.pdf,.png}
\usepackage{bigstrut}

\usepackage{csquotes}

\usepackage{algorithm, setspace}
\usepackage{algpseudocode}
\usepackage{url}
\usepackage{multirow}
\usepackage{float}
\usepackage{hyperref}

\title{\LARGE \bf
Controlling Assistive Robots with Learned Latent Actions
}

\author{Dylan P. Losey$^{1}$, Krishnan Srinivasan$^{1}$, Ajay Mandlekar$^{1}$, Animesh Garg$^{2}$, Dorsa Sadigh$^{1}$
\thanks{$^{1}$Stanford Intelligent and Interactive Autonomous Systems Group (ILIAD), Dept of Computer Science, Stanford University, Stanford, CA 94305. \newline
$^{2}$Animesh Garg is with University of Toronto, Vector Institute, and Nvidia. \newline
{(e-mail: dlosey@stanford.edu)}}%
}

\begin{document}
\maketitle

\begin{abstract} 

Assistive robotic arms enable users with physical disabilities to perform everyday tasks without relying on a caregiver. Unfortunately, the very dexterity that makes these arms useful also makes them challenging to teleoperate: the robot has more degrees-of-freedom than the human can directly coordinate with a handheld joystick. Our insight is that we can make assistive robots \textit{easier} for humans to control by leveraging \textit{latent actions}. Latent actions provide a low-dimensional embedding of high-dimensional robot behavior: for example, one latent dimension might guide the assistive arm along a pouring motion. In this paper, we design a teleoperation algorithm for assistive robots that \textit{learns} {latent actions} from task demonstrations. We formulate the controllability, consistency, and scaling properties that user-friendly latent actions should have, and evaluate how different low-dimensional embeddings capture these properties. Finally, we conduct two user studies on a robotic arm to compare our latent action approach to both state-of-the-art shared autonomy baselines and {a teleoperation strategy currently used by assistive arms}. Participants completed assistive eating and cooking tasks more efficiently when leveraging our latent actions, and also subjectively reported that latent actions made the task easier to perform. The video accompanying this paper can be found at: \href{https://youtu.be/wjnhrzugBj4}{\color{blue}{https://youtu.be/wjnhrzugBj4}}.

\end{abstract}

\smallskip



\section{Introduction}

For the nearly one million American adults that need assistance when eating, taking a bite of food or pouring a glass of water presents a significant challenge \cite{taylor2018americans}. Wheelchair-mounted robotic arms and other physically assistive robotic devices provide highly dexterous tools for performing these tasks without relying on help from a caregiver. In order to be effective, however, these assistive robots must be \textit{easily controllable} by their users.

Consider a person controlling a robotic arm to pour water into a glass. {Because of the physical limitations of their users, today's assistive arms utilize \textit{low-dimensional} control interfaces, such as joysticks \cite{gopinath2016human}. But the robot arm is \textit{high-dimensional}: it has many degrees-of-freedom (DoFs), and the human must precisely coordinate all of these interconnected DoFs to pour the water without spilling. In practice, user studies have demonstrated that controlling assistive arms is quite challenging due to the unintuitive mapping from low-DoF human inputs to high-DoF robot actions \cite{herlant2016assistive, argall2018autonomy}}.

\begin{figure*}[t]
    \vspace{0.5em}
	\begin{center}
		\includegraphics[width=2.0\columnwidth]{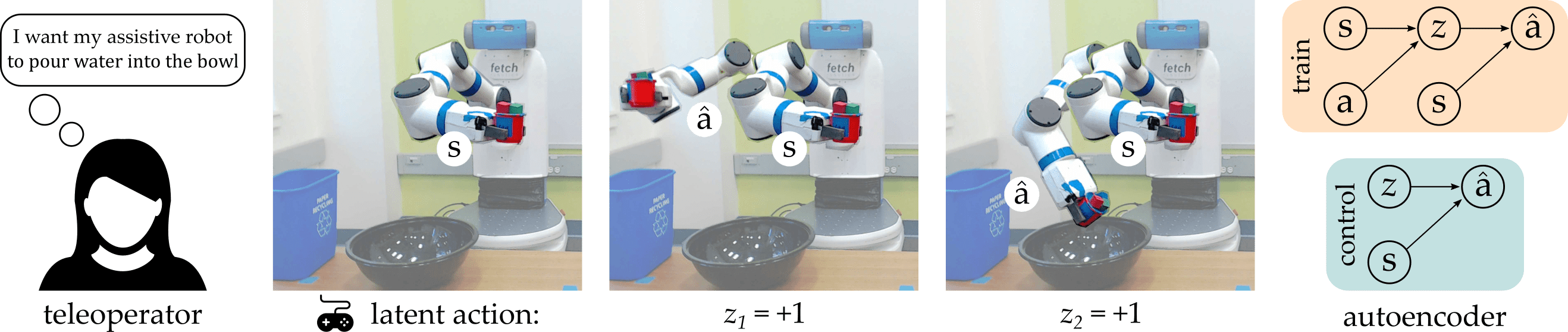}

		\caption{Human teleoperating a robot arm using latent actions. Because the robot has more DoFs than the human can directly control, we leverage low-DoF embeddings to learn a latent space $z$ for the user to intuitively interact with. Here the robot has been trained on demonstrations of pouring tasks, and learns a 2-DoF latent space. The first latent dimension $z_1$ moves the cup level with the table, and the second latent dimension $z_2$ tilts the cup. We explore how conditional autoencoders---such as the one shown on right---can be leveraged to learn these intuitive and human-controllable latent actions. {We train an encoder that finds our low-DoF embedding $z$ given the state $s$ and high-DoF action $a$. The decoder then recovers a high-DoF action $\hat{a}$ based on $z$ and $s$. During control, the low-DoF human input $z$ is enough to reconstruct their intended high-DoF, continuous robot action $\hat{a}$ conditioned on the current state $s$.}}

		\label{fig:front}
	\end{center}
	
	\vspace{-2em}
	
\end{figure*}

Current approaches solve this problem when the human's goals are \textit{discrete}: e.g., when the robot should pour water into either glass A or glass B. By contrast, we here propose an approach for controlling the robot in \textit{continuous} spaces using low-DoF actions learned from data. Our insight is that:
\begin{center}
\emph{High-DoF robot actions can often be} embedded \emph{into intuitive, human-controllable, and low-DoF latent actions}
\end{center}
{Latent actions are a low-DoF representation that captures the most important aspects of high-DoF actions}. Returning to our pouring example: the human wants the robot arm to (a) carry the cup level with the table and (b) perform a pouring action. Intuitively, this should be reflected in the {latent actions}: one {latent dimension} should cause the robot to move the cup along the table, while the other should make the robot pour more or less water (see Fig.~\ref{fig:front}).

We explore methods for \textit{learning} these low-DoF latent actions from task-specific training data. We envision settings where the robot has access to demonstrations of related tasks (potentially provided by the caregiver), and the user---in an online setting---wants to control the robot to perform a new task: e.g., now the cup is located in a different place, and the person only wants half a glass of water. In practice, we find that some models result in expressive and intuitive latent actions, and that users can control robots equipped with these models to complete eating and cooking tasks.

Overall, we make the following contributions:

\noindent\textbf{Formalizing Desirable Properties of Latent Actions.} We formally specify a list of properties that user-friendly latent actions should satisfy. This includes \emph{controllability}, i.e., there must be a sequence of latent actions that move the robot to the desired state, and \emph{consistency}, i.e., the robot should {always behave similarly under a given} learned latent action.

\noindent\textbf{Learning Latent Actions through Autoencoders.}
We learn latent actions using different autoencoder models, and {compare how these models perform with respect to our desired properties}. We find that {models which are conditioned on the robot's current state} accurately reconstruct high-DoF actions from human-controllable, low-DoF inputs.

\noindent\textbf{Evaluating Latent Actions with User Studies.} We implement our approach on a robot arm, and compare to state-of-the-art shared autonomy baselines and {a current control strategy for assistive robotic arms}. We find that---during eating and cooking user studies---learned latent actions resulted in improved objective and subjective performance.

\smallskip

\section{Related Work}

In this paper we leverage learning techniques to identify low-DoF latent actions for teleoperating high-DoF robots. {Prior works on shared autonomy have addressed (a) using predefined mappings from low-DoF human inputs to robot actions, and (b) learning mappings when the human's action space has the same number of DoFs as the robot. Previous research on latent representations (c) focused on autonomous robots that are acting without a human in-the-loop.}

\smallskip

\noindent\textbf{Shared Autonomy.} Under shared autonomy, the robot combines user inputs with autonomous assistance: {this method has been applied to settings where the human's input is low-dimensional} \cite{argall2018autonomy, carlson2013brain, muelling2015autonomy}. Recent works explored tasks where the human wants their assistive arm to reach a goal (e.g., pick up a cup) \cite{javdani2018shared, dragan2013policy, gopinath2016human, aronson2018eye, broad2018learning}. Here the robot maintains a belief over possible goals, and increasingly assists the human as it becomes confident about their intended goal \cite{dragan2013policy, javdani2018shared}. {We point out that---similar to \cite{abi2018user}---the mapping between the human's low-DoF input and the robot's high-DoF action is pre-defined, and not learned from data.}

{Other shared autonomy works consider settings where the human's input has the same number of DoFs as the robot's action space \cite{losey2018review, rakita2017motion, abi2017learning, reddy2018shared}.} Our paper is most related to shared autonomy research by Reddy \textit{et al.} \cite{reddy2018shared}, where the robot learns a mapping between humans inputs and their intended actions using reinforcement learning; however, in \cite{reddy2018shared} the human inputs are the same dimension as the robot action, and thus {there is no need to learn an embedding for control.}

\smallskip

\noindent\textbf{Learning Latent Representations.} To identify low-DoF embeddings of complex state-space models, we turn to works that learn latent representations from data. Recent research has learned latent dynamics \cite{watter2015embed}, trajectories \cite{reyes2018self}, plans \cite{lynch2019learning}, policies \cite{edwards2019imitating}, {and skills for reinforcement learning \cite{hausman2018learning}}. These methods typically leverage autoencoder models \cite{kingma2013auto}, {but do not include a human in-the-loop or teleoperation.}

Here, we leverage autoencoders to learn a consistent and controllable latent representation for assistive robotics. Previous teleoperation literature has explored principal component analysis (PCA) for reducing the user's input dimension \cite{artemiadis2010emg, ciocarlie2009hand}. We will compare our method to this PCA baseline.
\section{Problem Statement}

We formulate a task as a discrete-time Markov Decision Process (MDP) $\mathcal{M} = (\mathcal{S}, \mathcal{A}, \mathcal{T}, R, \gamma, \rho_0)$. {Here $s \in \mathcal{S} \subseteq \mathbb{R}^n$ is the state space, $a \in \mathcal{A} \subseteq \mathbb{R}^m$ is the action space}, $\mathcal{T}(s, a)$ is the transition function, $R(s) = \mathbbm{1}\{\text{task is solved in } s\}$ is a sparse reward function that indicates task success, $\gamma \in [0, 1)$ is the discount factor, and $\rho_0(\cdot)$ is the initial state distribution. We assume access to a dataset of task demonstrations, and want to learn the latent action space by leveraging this dataset. Formally, we have a dataset of state-action pairs $\mathcal{D} = \{(s_0, a_0), (s_1, a_1), \ldots \}$, and seek to learn a latent action space $\mathcal{Z} \subset \mathbb{R}^d$ that is of lower dimension than the original action space ($d < m$), along with a function $\phi : \mathcal{Z} \times \mathcal{S} \mapsto \mathcal{A}$ that maps latent actions to robot actions.

Recall our motivating example, where the human is leveraging latent actions to make their assistive robot pour water. There are several \textit{properties} that the human expects latent actions to have: e.g., the human should be able to guide the robot by smoothly changing the joystick direction, and the robot should never abruptly become more sensitive to the human's inputs. In what follows, we \textit{formalize} the properties that make latent actions intuitive. These properties will guide our approach, and provide a principled way of assessing the usefulness of latent actions with humans in-the-loop.

\smallskip

\noindent\textbf{Latent Controllability.} Let $s_i, s_j \in \mathcal{D}$ be two states from the dataset of demonstrations, and let $s_{1}, s_{2}, ..., s_{K}$ be the sequence of states that the robot visits when starting in state $s_0 = s_i$ and taking latent actions $z_1, ..., z_K$. The robot transitions between the visited states using the learned latent space: $s_{k} = \mathcal{T}(s_{k-1}, \phi(z_{k-1}, s_{k-1}))$. Formally, we say that a latent action space $\mathcal{Z}$ is \textit{controllable} if for every such pairs of states $(s_i, s_j)$ there exists a sequence of latent actions $\{z_k\}_{k=1}^K, z_k \in \mathcal{Z}$ such that $s_j = s_K$. In other words, a latent action space is controllable if it can move the robot between pairs of start and goal states from the dataset.

\smallskip

\noindent\textbf{Latent Consistency.} We define a latent action space $\mathcal{Z}$ as \textit{consistent} if the same latent action $z \in \mathcal{Z}$ has a similar effect on how the robot behaves in nearby states. We formulate this similarity via a task-dependent metric $d_M$: e.g., in pouring tasks $d_M$ could measure the orientation of the robot's end-effector. Applying this metric, consistent latent actions should satisfy:  $d_M(\mathcal{T}(s_1,\phi(z, s_1)),\mathcal{T}(s_2,\phi(z, s_2))) < \epsilon$ for $\|s_1 - s_2\| < \delta$ for some $\epsilon, \delta>0$.

\smallskip

\noindent\textbf{Latent Scaling.} Finally, a latent action space $\mathcal{Z}$ is \textit{scalable} if applying larger latent actions leads to larger changes in state. In other words, we would like $\lVert s-s'\rVert\to \infty$ as $\lVert z \rVert \to \infty$, where $s'=\mathcal{T}(s, \phi(z, s))$.
\section{Methods}

Now that we have formally introduced the properties that a user-friendly latent space should satisfy, we will explore low-DoF embeddings that capture these properties. We are interested in models that balance \textit{expressiveness} with \textit{intuition}: the embedding must reconstruct high-DoF actions while remaining controllable, consistent, and scalable. We assert that only models which reason over the robot's state when decoding the human's inputs can accurately and intuitively interpret the latent action.

\subsection{Models}

\noindent\textbf{Reconstructing Actions.} Let us return to our pouring example: when the person applies a low-DoF joystick input, the robot completes a high-DoF action. We use \textit{autoencoders} to move between these low- and high-DoF action spaces. Define $\psi : \mathcal{S} \times \mathcal{A} \rightarrow \mathcal{Z}$ as an \textit{encoder} that embeds the robot's behavior into a latent space, and define $\phi' : \mathcal{Z} \rightarrow \mathcal{A}$ as a \textit{decoder} that reconstructs a high-DoF robot action {$\hat{a} \in \mathcal{A}$} from this latent space (see Fig.~\ref{fig:front}). To encourage models to learn latent actions that accurately reconstruct high-DoF robot behavior, we incorporate the reconstruction error $\|a - \hat{a}\|^2$ into the model's loss function, {which measures the difference between the demonstrated action $a$ and the model's estimate $\hat{a}$}.  

\smallskip

\noindent\textbf{Regularizing Latent Actions.} When the user slightly tilts the joystick, the robot should not suddenly pour its entire glass of water. To better ensure this consistency and scalability, we incorporate a \textit{normalization} term into the model's loss function. Let us define $\psi : \mathcal{S} \times \mathcal{A} \rightarrow \mathbb{R}^d \times \mathbb{R}_+^d$ as an encoder that outputs the mean $\mu$ and covariance $\sigma$ of the latent action space. We penalize the divergence between this latent action space and a normal distribution: $KL(\mathcal{N}(\mu, \sigma) ~\|~ \mathcal{N}(0, 1))$. Variational autoencoder (VAE) models trade-off between this normalization term and reconstruction error \cite{kingma2013auto}.

\smallskip

\noindent\textbf{Conditioning on State.} Importantly, we recognize that the \textit{meaning} of the human's joystick input often depends on what the robot is doing. When the robot is holding a glass, pressing down on the joystick indicates that the robot should pour water; but---when the robot's gripper is empty---it does not make sense for the robot to pour! So that robots can associate meanings with latent actions, we \textit{condition} the interpretation of the latent action on the robot's current state. Define $\phi : \mathcal{Z} \times \mathcal{S} \rightarrow \mathcal{A}$ as a decoder that now makes decisions based on both $z$ and $s$. We expect that conditional autoencoders (cAE) and conditional variational autoencoders (cVAE) which use $\phi$ will learn more expressive and controllable actions than their non-state conditioned counterparts.

\subsection{Algorithm}

\begin{algorithm}[b]
\caption{Learning Latent Control for Assistive Robots}
  \label{deltaco}
\begin{algorithmic}[1]
    \State Collect dataset $\mathcal{D} = \{(s_0,a_0), (s_1,a_1), \ldots \}$ from kinesthetic demonstrations
    \State Train autoencoder to minimize loss $L(s,a)$ on $\mathcal{D}$
    \State Align the learned latent space
    \For{$t \gets 1, 2, \ldots, T$}
        \State Set latent action $z_t$ as human's joystick input
        \State Execute reconstructed action $\hat{a}_t\gets \phi(z_t, s_t)$
    \EndFor
\end{algorithmic}
\end{algorithm}

Our approach for training and leveraging these models is shown in Algorithm~\ref{deltaco}. First, the robot obtains demonstrations of related tasks---these could be provided by a caregiver, or even collected from another end-user. The robot then trains a low-dimensional embedding using this data and one of the models described above. We manually align the learned latent dimensions with the joystick DoF; for example, we rotate $z$ so that up/down on the joystick corresponds to the latent DoF that pours/straightens the glass. When the user interacts with the robot, their inputs are treated as $z$, and the robot utilizes its decoder $\phi$ to reconstruct high-DoF actions.
\section{Simulations}

To test if the proposed low-DoF embeddings capture our desired user-friendly properties, we perform simulations on robot arms. The simulated robots have more DoF than needed to complete the task, and thus must learn to coordinate their redundant joints when decoding human inputs.

\subsection{Setup}

We simulate one- and two-arm planar robots, where each arm has five revolute joints and links of equal length. The state $s \in \mathbb{R}^{n}$ is the robot's joint position, and the action $a \in \mathbb{R}^{n}$ is the robot's joint velocity. Hence, the robot transitions according to: $s_{t+1} = s_t + a_t \cdot dt$, where $dt$ is the step size. Demonstrations consist of trajectories of state-action pairs: in each of different simulated tasks, the robot trains with a total of $10000$ state-action pairs.

\smallskip

\noindent \textbf{Tasks.} We consider four different tasks.
\begin{enumerate}
    \item \textit{Sine}: a single 5-DoF robot arm moves its end-effector along a sine wave with a 1-DoF latent action
    \item \textit{Rotate}: two robot arms are holding a box, and rotate that box about a fixed point using a 1-DoF latent action
    \item \textit{Circle}: one robot arm moves along circles of different radii with a 2-DoF latent action
    \item \textit{Reach}: a one-arm robot reaches from a start location to a goal region with a 1-DoF latent action
\end{enumerate}

\smallskip

\noindent \textbf{Model Details.} We examine models such as PCA, AE, VAE, and state conditioned models such as cAE and cVAE. The encoders and decoders contain between two and four linear layers (depending on the task) with a $tanh(\cdot)$ activation function. The loss function is optimized using Adam with a learning rate of $1e^{-2}$. Within the VAE and cVAE, we set the normalization weight $<1$ to avoid posterior collapse.

\smallskip

\noindent\textbf{Dependent Measures.} To determine \textit{accuracy}, we measure the mean-squared error between the intended actions $a$ and reconstructed actions $\hat{a}$ on a test set of state-action pairs $(s,a)$ drawn from the same distribution as the training set. 

To test model \textit{controllability}, we select pairs of start and goal states $(s_i,s_j)$ from the test set, and solve for the latent actions $z$ that minimize the error between the robot's current state and $s_j$. We then report this minimum state error.

We jointly measure \textit{consistency} and \textit{scalability}: to do this, we select $25$ states along the task, and apply a fixed grid of latent actions $z_i$ from $[-1, +1]$ at each state. For every $(s,z)$ pair we record the distance and direction that the end-effector travels (e.g., the direction is $+1$ if the end-effector moves right). We then find the best-fit line relating $z$ to distance times direction, and report its $R^2$ error.

Our results are averaged across $10$ trained models of the same type, and are listed in the form $mean \pm SD$.

\smallskip

\noindent \textbf{Hypotheses.} We have the following two hypotheses:
\begin{displayquote}
\textbf{H1.} \textit{Only models conditioned on the state will accurately reconstruct actions from low-DoF inputs.}
\end{displayquote}
\begin{displayquote}
\textbf{H2.} \textit{State conditioned models will learn a latent space that is controllable, consistent, and scalable.}
\end{displayquote}

\subsection{Sine Task}

\begin{figure}[t]
\vspace{0.45em}
	\begin{center}
		\includegraphics[width=1.0\columnwidth]{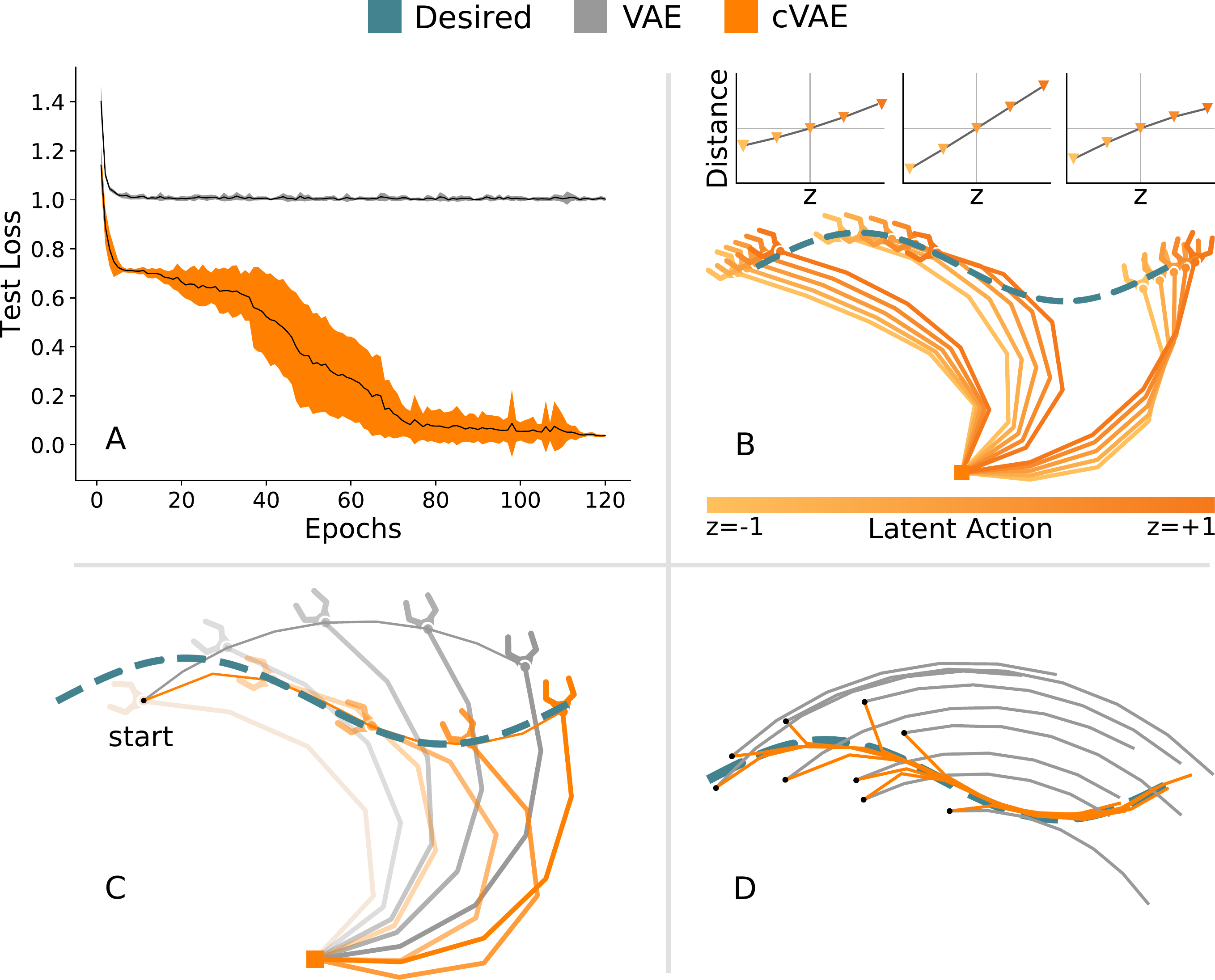}

		\caption{Results for the \textit{Sine} task. (A) mean-squared error between intended and reconstructed actions normalized by PCA test loss. (B) effect of the latent action $z$ at three states along the sine wave for the cVAE model. Darker colors correspond to $z>0$ and lighter colors signify $z<0$. Above we plot the distance that the end effector moves along the sine wave as a function of $z$ at each state. (C) rollout of robot behavior when applying a constant latent input $z=+1$, where both VAE and cVAE start at the same state. (D) end-effector trajectories for multiple rollouts of VAE and cVAE.}

		\label{fig:sine}
	\end{center}
	
	\vspace{-2em}
	
\end{figure}

This task and our results are shown in Fig.~\ref{fig:sine}. We find that including state conditioning greatly improves \textit{accuracy} when compared to the PCA baseline: AE and VAE incur $98.0\pm 0.6\%$ and $100\pm0.8\%$ of the PCA loss, while cAE and cVAE obtain $1.37\pm1.2\%$ and $3.74\pm0.4\%$ of the PCA loss, respectively. {We likewise observe that the state conditioned models are more \textit{controllable} than their alternatives. When using the learned latent actions to move between $1000$ randomly selected pairs of states along the sine wave, cAE and cVAE are on average $5.6\%$ and $11.1\%$ as far from the goal as PCA. By contrast, models without state conditioning (i.e., AE and VAE) performed worse than the PCA baseline, with $104\%$ error and $106\%$ error.}

When evaluating \textit{consistency} and \textit{scalability}, we discover that every model's relationship between latent actions and robot behavior can be modeled as approximately linear: PCA has the highest $R^2 = 0.99$, while cAE and cVAE have the lowest $R^2 = 0.94 \pm 0.04$ and $R^2 = 0.95 \pm 0.01$.

\subsection{Rotate Task}

\begin{figure}[t]
\vspace{0.45em}
	\begin{center}
		\includegraphics[width=1.0\columnwidth]{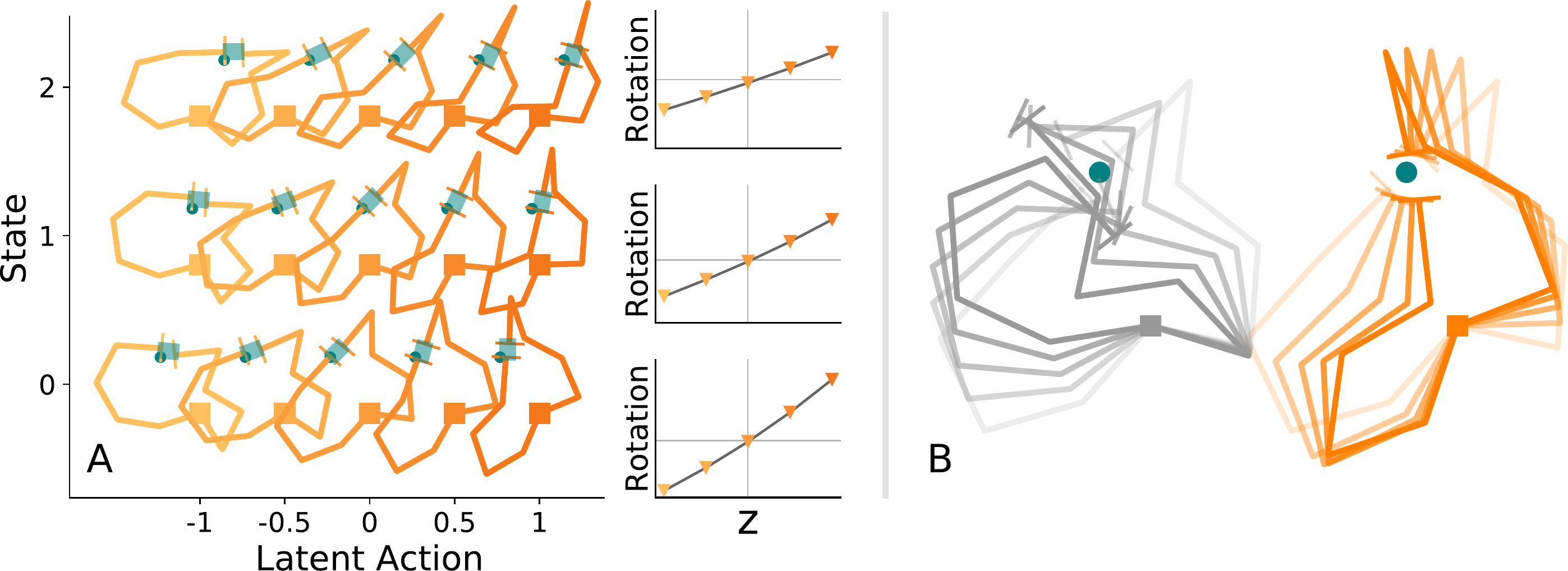}

		\caption{Results for the \textit{Rotate} task. (A) the robot uses two arms to hold a light blue box, and learns to rotate this box around the fixed point shown in teal. Each state corresponds to a different fixed point, and positive $z$ causes counterclockwise rotation. On right we show how $z$ affects the rotation of the box at each state. (B) rollout of the robot's trajectory when the user applies $z=+1$ for VAE and cVAE models, where both models start in the same state. Unlike the VAE, the cVAE model coordinates its two arms.}

		\label{fig:rotate}
	\end{center}
		
	\vspace{-2em}
	
\end{figure}

We summarize the results for this two-arm task in Fig.~\ref{fig:rotate}. Like in the \textit{Sine} task, the models conditioned on the current state are more \textit{accurate} than their non-conditioned counterparts: AE and VAE have $28.7\pm4.8\%$ and $38.0\pm5.8\%$ of the PCA baseline loss, while cAE and cVAE reduce this to $0.65\pm0.05\%$ and $0.84\pm0.07\%$. The state conditioned models are also more \textit{controllable}: when using the learned $z$ to rotate the box, AE and VAE have $56.8\pm9\%$ and $71.5\pm8\%$ as much end-effector error as the PCA baseline, whereas cAE and cVAE achieve $5.4\pm0.1\%$ and $5.9\pm0.1\%$ error.

When testing for \textit{consistency} and \textit{scalability}, we measure the relationship between the latent action $z$ and the change in orientation for the end-effectors of both arms (i.e., ignoring their location). Each model exhibits a linear relationship between $z$ and orientation: $R^2 = 0.995\pm0.004$ for cVAE and $R^2 = 0.996\pm0.002$ for cVAE. In other words, there is an approximately linear mapping between $z$ and the orientation of the box that the two arms are holding.

\subsection{Circle Task}

\begin{figure}[t]
\vspace{0.45em}
	\begin{center}
		\includegraphics[width=1.0\columnwidth]{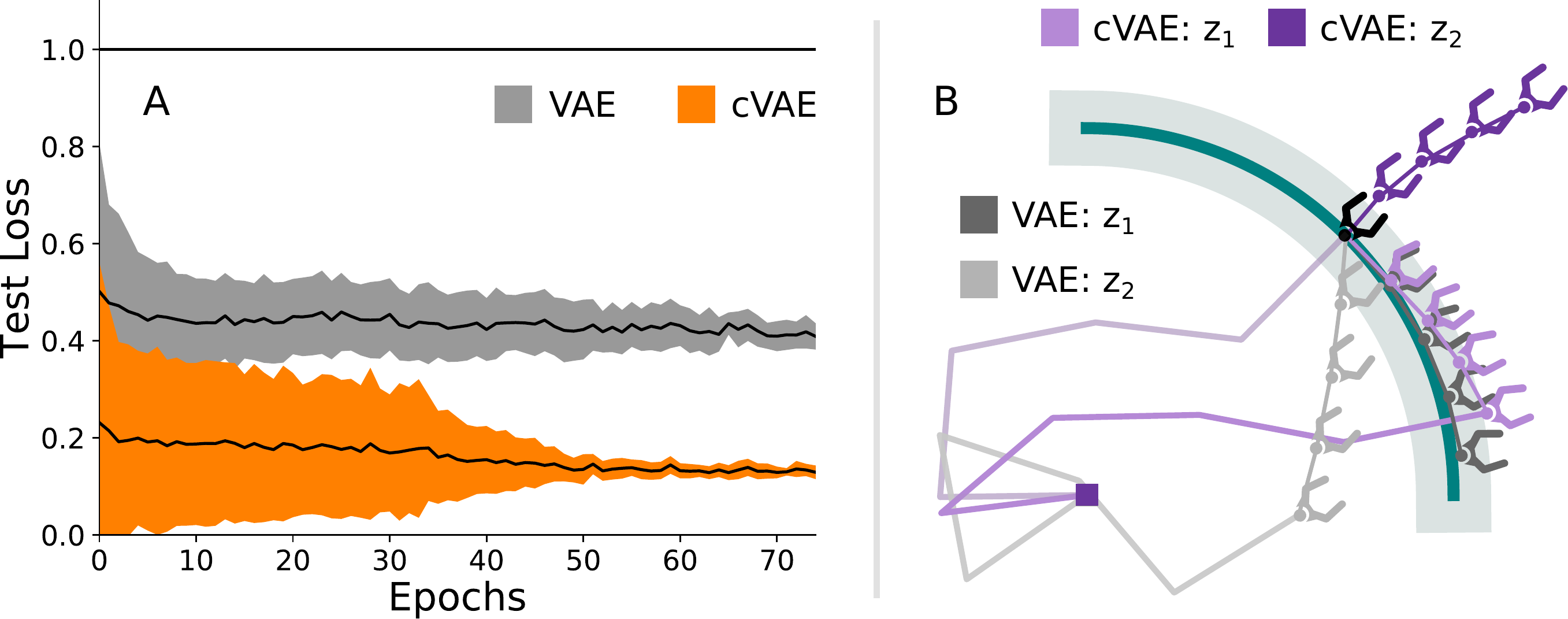}

		\caption{Results for the \textit{Circle} task. (A) mean-squared error between desired and reconstructed actions normalized by the PCA test loss. (B) 2-DoF latent action space $z = [z_1, z_2]$ for VAE and cVAE models. The current end-effector position is shown in black, and the colored grippers depict how changing $z_1$ or $z_2$ affects the robot's state. Under the cVAE model, these latent dimensions move the end-effector tangent or orthogonal to the circle.}

		\label{fig:circle}
	\end{center}
	
	\vspace{-0.5em}
	
\end{figure}

Next, consider the one-arm task in Fig.~\ref{fig:circle} where the robot has a 2-DoF latent action space. We here focus on the learned latent dimensions $z = [z_1,z_2]$, and examine how these latent dimensions correspond to the underlying task. Recall that the training data consists of state-action pairs which translate the robot's end-effector along (and between) circles of different radii. Ideally, the learned latent dimensions correspond to these axes, e.g., $z_1$ controls tangential motion while $z_2$ controls orthogonal motion. Interestingly, we found that this intuitive mapping is \textit{only} captured by the state conditioned models. The average angle between the directions that the end-effector moves for $z_1$ and $z_2$ is $27\pm 20^{\circ}$ and $34 \pm 15^{\circ}$ for AE and VAE models, but this angle increases to $72 \pm 9^{\circ}$ and $74 \pm 12^{\circ}$ for the cAE and cVAE (ideally $90^{\circ}$). The state conditioned models better \textit{disentangle} their low-dimensional embeddings, supporting our hypotheses and demonstrating how these models produce user-friendly latent spaces.

\subsection{Reach Task}

\begin{figure}[t]
\vspace{0.25em}
	\begin{center}
		\includegraphics[width=1.0\columnwidth]{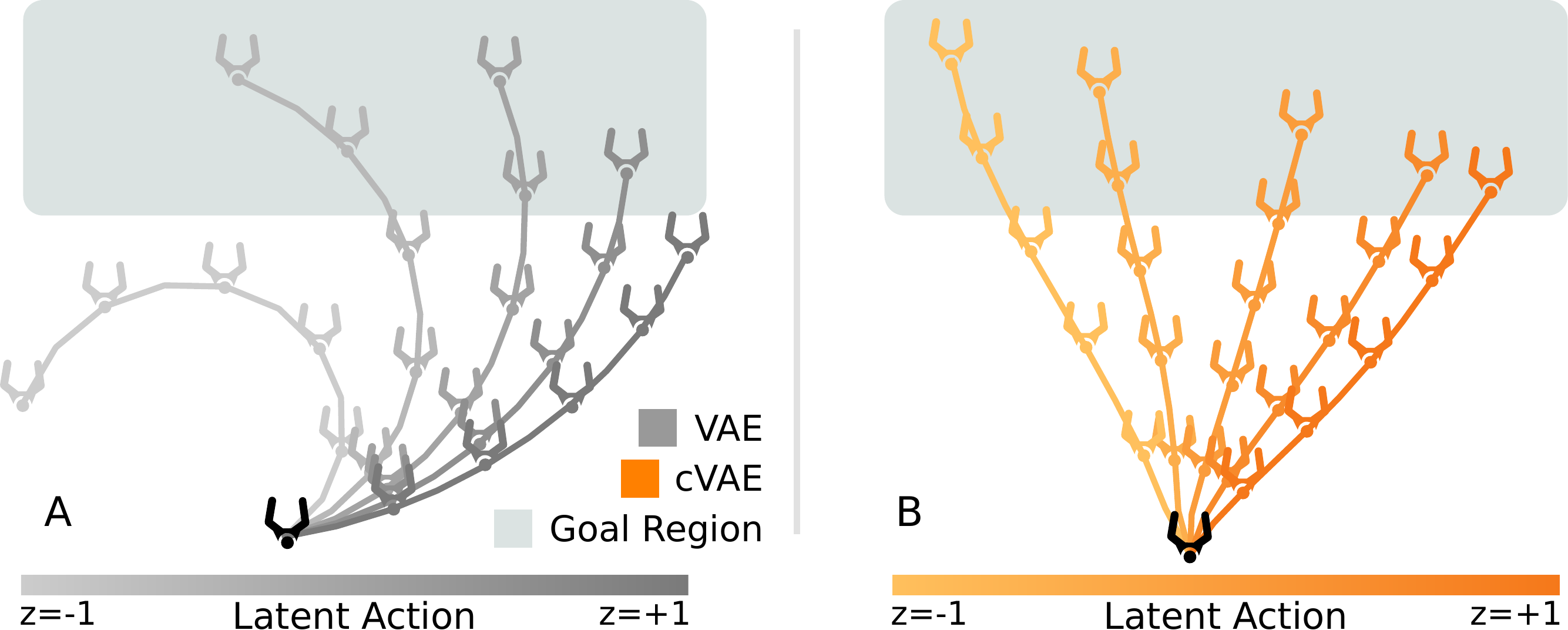}

		\caption{Results for the \textit{Reach} task. In both plots, we show the end-effector trajectory when applying constant inputs $z \in [-1,+1]$. The lightest color corresponds to $z=-1$ and the darkest color is $z=+1$. The goal region is highlighted, and the initial end-effector position is black. (A) trajectories with the VAE model. (B) trajectories with the cVAE model. The latent action $z$ controls which part of the goal region the trajectory moves towards.}

		\label{fig:reach}
	\end{center}
	
	\vspace{-2em}
	
\end{figure}

In the final task, a one-arm robot trains on trajectories that move towards a goal region (see Fig.~\ref{fig:reach}). The robot learns a 1-DoF latent space, where $z$ controls the direction that the trajectory moves (i.e., to the left or right of the goal region). We focus on \textit{controllability}: can robots utilize latent actions to reach their desired goal? In order to test controllability, we sample $100$ goals randomly from the goal region, and compare robots that attempt to reach these goals with either VAE or cVAE latent spaces. The cVAE robot more accurately reaches its goal: the $L_2$ distance between the goal and the robot's final end-effector position is $0.57\pm0.38$ under VAE and $0.48\pm0.5$ with cVAE. Importantly, using state conditioning improves not only the movement accuracy but also the movement \textit{quality}. The average start-to-goal trajectory is $5.1\pm2.8$ units when using the VAE, and this length drops to $3.1\pm0.5$ with the cVAE model.

\smallskip

\noindent\textbf{Summary.} Viewed together, the results of our \textit{Sine}, \textit{Rotate}, \textit{Circle}, and \textit{Reach} tasks support hypotheses H1 and H2. The state conditioned models more \textit{accurately} reconstruct high-DoF actions from low-DoF embeddings (H1), and also exhibit the user-friendly properties of \textit{controllability}, \textit{consistency}, and \textit{scalability} (H2). We also observed that the latent dimensions naturally aligned themselves with the underlying DoF of the task, and that we could leverage the state conditioned models to embed robot trajectories.
\section{User Studies} \label{sec:userstudy}

To evaluate whether actual humans can use learned latent actions to teleoperate robots and perform everyday tasks, we conducted two user studies on a 7-DoF robotic arm (Fetch, Fetch Robotics). In the first study, we compared our proposed approach to state-of-the-art shared autonomy methods when the robot has a \textit{discrete} set of possible goals. In the second study, participants completed a cooking task with {\textit{continuous} goal spaces} using either {a teleoperation method currently employed by assistive arms} or our learned latent actions. For both studies the participants controlled the robot arm with a low-DoF teleoperation interface (a handheld joystick).

\begin{figure}[t]
	\vspace{0.45em}

	\begin{center}
		\includegraphics[width=1.0\columnwidth]{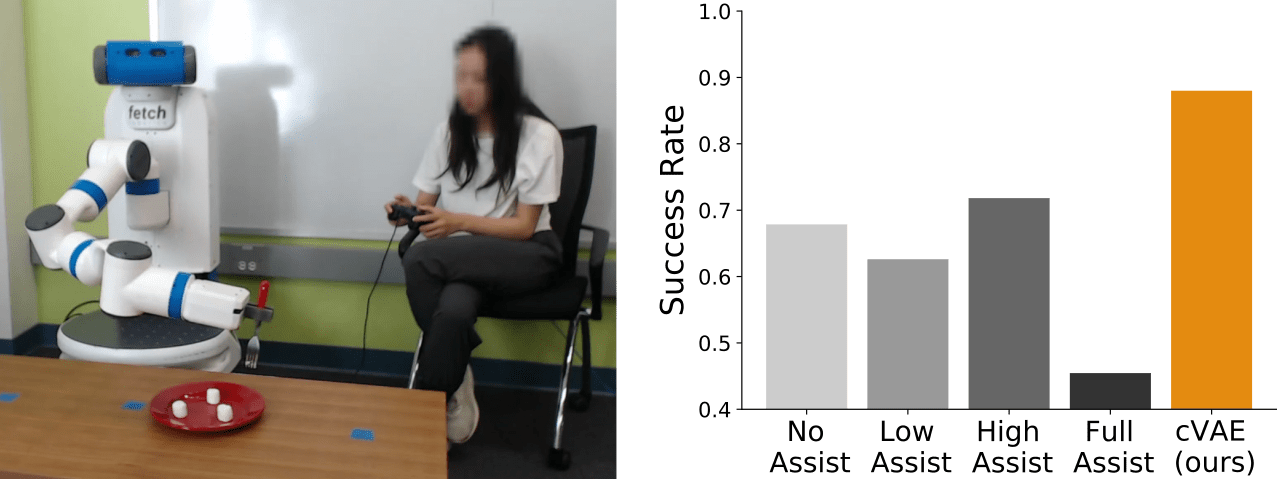}

		\caption{Experimental setup for our first user study. (Left) in this eating task, the participant uses a two-DoF joystick to guide the robot to pick up their desired morsel from a discrete set. (Right) we compare our latent action approach to shared autonomy {baselines} from the HARMONIC dataset.}

		\label{fig:harmonic1}
	\end{center}
	
	\vspace{-2em}
	
\end{figure}

\subsection{{Discrete Goal Space}: Latent Actions vs. Shared Autonomy}

In our first user study we implemented the assistive eating task from the HARMONIC dataset \cite{NewmanHARMONIC2018} (see Fig.~\ref{fig:harmonic1}). Here the human is guiding the robot to pick up a bite of food. There are three morsels near the robot---i.e., three possible goals---and the human wants the robot to reach one of these discrete goals. The HARMONIC dataset reports the performance of $24$ people who completed this task with different levels of shared autonomy \cite{javdani2018shared}. Under shared autonomy, the robot infers from the human's inputs which goal they are trying to reach, and then provides assistance towards that goal. We here conduct an additional experiment to compare our latent action method to this {set of baselines}.

\smallskip

\noindent\textbf{Independent Variables.} We manipulated the robot's teleoperation strategy with five levels: the four conditions from the HARMONIC dataset plus our proposed cVAE method. In the first four conditions, the robot provided no assistance (\textit{No Assist}), or interpolated between the human's input and an assistive action (\textit{Low Assist}, \textit{High Assist}, and \textit{Full Assist}). \textit{High Assist} was the most effective strategy from this group: when interpolating, here the assistive action was given twice the weight of the human's input. In our \textit{cVAE} approach, the human's joystick inputs were treated as latent actions $z$ (and the robot provided no other assistance). We trained our cVAE model on demonstrations from the HARMONIC dataset.

\smallskip

\noindent\textbf{Dependent Measures.} We measured the fraction of trials in which the robot picked up the correct morsel of food (\textit{Success Rate}), the amount of time needed to complete the task (\textit{Completion Time}), the total magnitude of the human's input (\textit{Joystick Input}), and the distance traveled by the robot's end-effector (\textit{Trajectory Length}).

\smallskip

\noindent\textbf{Hypothesis.} We had the following hypothesis: 
\begin{displayquote}
\textbf{H3.} \emph{Teleoperating with learned latent actions will improve task success while reducing the completion time, joystick inputs, and trajectory length.}
\end{displayquote}

\smallskip

\noindent\textbf{Experimental Setup.} Participants interacted with a handheld joystick while watching the robotic arm. The robot held a fork; during the task, users teleoperated the robot to position this fork directly above their desired morsel. We selected the robot's start state, goal locations, and movement speed to be consistent with the HARMONIC dataset.

\smallskip

\noindent\textbf{Participants and Procedure.} Our participant pool consisted of ten Stanford University affiliates who provided informed consent ($3$ female, average participant age $23.9\pm2.8$ years). Following the same protocol as the HARMONIC dataset, each participant was given up to five minutes to familiarize themselves with the task and joystick, and then completed five recorded trials using our \textit{cVAE} approach. At the start of each trial the participant indicated which morsel they wanted the robot to reach; the trial ended once the user pushed a button to indicate that the fork was above their intended morsel. We point out that participants only completed the task with the \textit{cVAE} condition; other teleoperation strategies are benchmarked in Newman \textit{et al.} \cite{NewmanHARMONIC2018}.

\smallskip 

\begin{figure}[t]
\vspace{0.25em}
	\begin{center}
		\includegraphics[width=1.0\columnwidth]{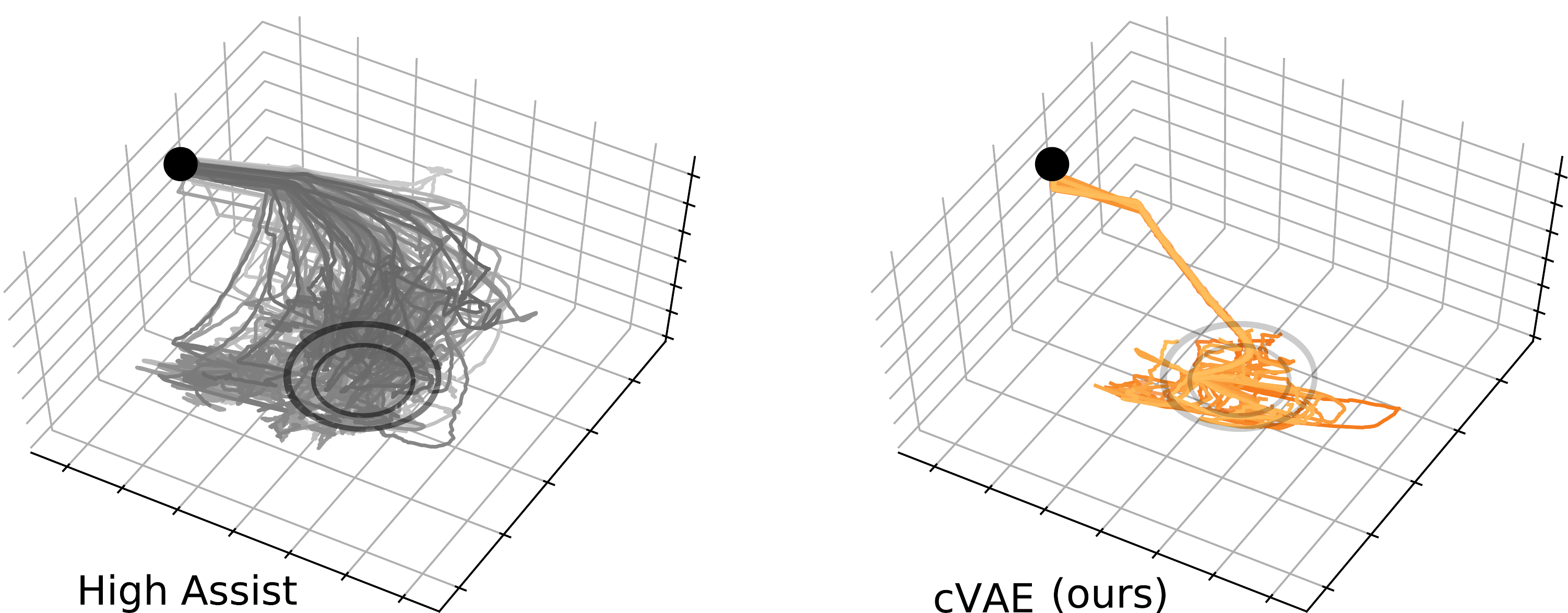}

		\caption{End-effector trajectories from \textit{High Assist} and \textit{cVAE} conditions. The robot starts at the black dot, and moves to position itself over the plate.}

		\label{fig:harmonic2}
	\end{center}
	
	\vspace{-0.5em}
	
\end{figure}

\begin{figure}[t]
	\begin{center}
		\includegraphics[width=1.0\columnwidth]{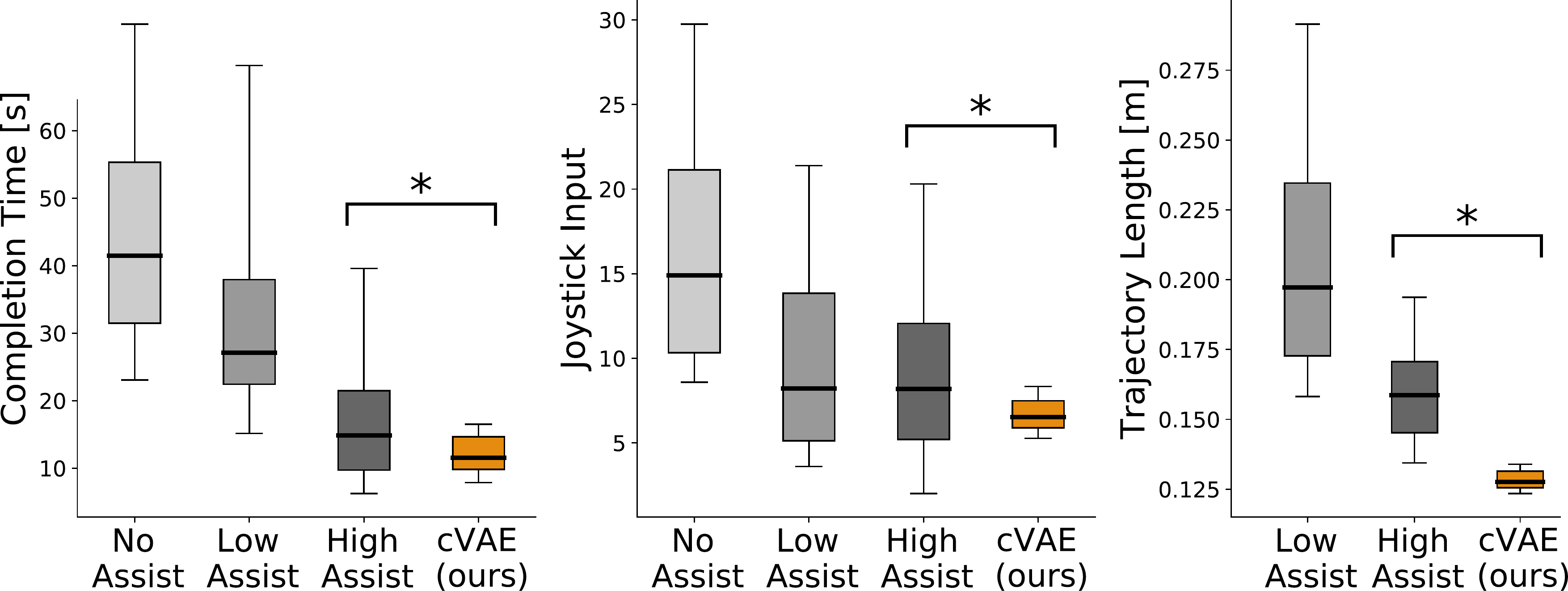}

		\caption{Objective results from the eating user study. We found that \textit{cVAE} led to faster task completion with less user input and end-effector motion. The \textit{Full Assist} condition performed worse than \textit{High Assist} across the board (omitted for clarity). Error bars show the $10$ and $90$ percentiles, and $*$ denotes statistical significance ($p < .05$).}
		
		\label{fig:harmonic3}
	\end{center}
	
	\vspace{-2em}
	
\end{figure}

\noindent\textbf{Results.} We display example robot trajectories in Fig.~\ref{fig:harmonic2} and report our dependent measures in Figs.~\ref{fig:harmonic1} and \ref{fig:harmonic3}. Inspecting these example trajectories, we observe that the \textit{cVAE} model learned latent actions that move the robot's end-effector into a region above the plate. Users controlling the robot with \textit{cVAE} reached their desired morsel in $44$ of the $50$ total trials, yielding a higher \textit{Success Rate} than the assistance baselines. To better compare \textit{cVAE} to the \textit{High Assist} condition, we performed independent t-tests. We found that participants that used the \textit{cVAE} model took statistically significant lower \textit{Completion Time} ($t(158) = 2.95$, $p < .05$), \textit{Joystick Input} ($t(158) = 2.49$, $p < .05$), and \textit{Trajectory Length} ($t(158) = 9.39$, $p < .001$), supporting our hypothesis H3.

\smallskip

\noindent\textbf{Summary.} Assistive robots that learn a mapping from low- to high-DoF actions can {complete tasks with discrete goals}. Users teleoperating the \textit{cVAE} robot reached their preferred goal more accurately than shared autonomy baselines, while requiring less time, effort, and movement.

\subsection{{Continuous Goal Space}: Latent Actions vs. End-Effector} \label{sec:cooking}

\begin{figure*}[t]
\vspace{0.45em}
	\begin{center}
		\includegraphics[width=1.4\columnwidth]{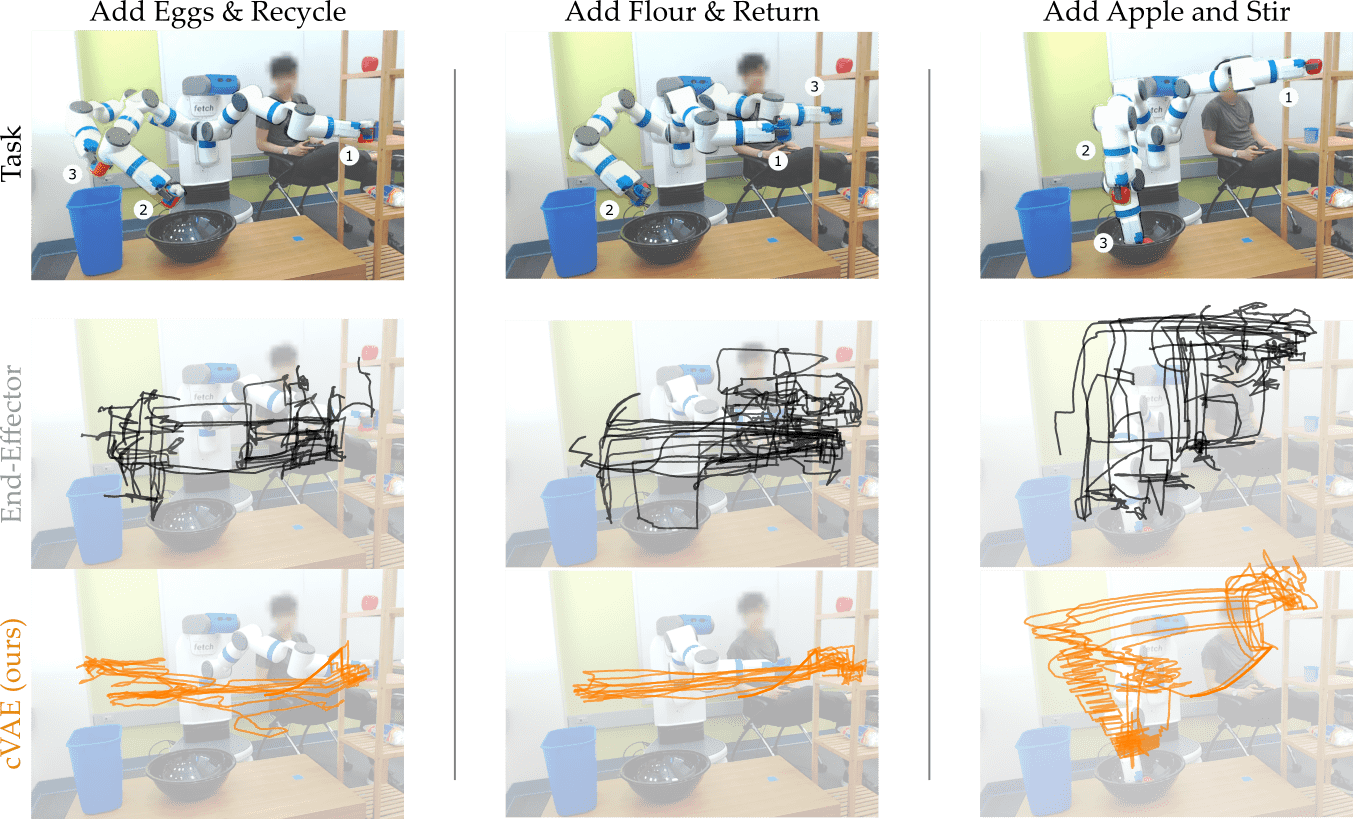}

		\caption{Setup for our second user study. (Top row) the participant is teleoperating an assistive robot to make their recipe. This recipe is broken down into three sub-tasks. On left the robot picks up eggs, pours them into the bowl, then drops the container into the recycling. In middle the robot picks up flour, pours it into the bowl, then returns the container to the shelf. On right the robot grasps an apple, places it in the bowl, then stirs the mixture. (Middle row) example robot trajectories when the person directly controls the robot's \textit{End-Effector}. (Bottom row) example trajectories when using \textit{cVAE} to learn latent actions. Comparing the example trajectories, we observe that \textit{cVAE} resulted in robot motions that more smoothly and directly accomplished the task.}

		\label{fig:cooking1}
	\end{center}
	
	\vspace{-2em}
	
\end{figure*}

{Real-world tasks often move beyond discrete goals: instead of reaching for an object that must either be in position A or position B, objects may lie in continuous regions (i.e., anywhere on a shelf)}. In our second user study, we therefore focus on a cooking scenario with {\textit{continuous} goals spaces} (see Fig.~\ref{fig:cooking1}). The user wants their assistive robot to help them make a recipe: this requires picking up ingredients from the shelf, pouring them into a bowl, recycling empty containers---or returning half-filled containers to the shelf---and then stirring the mixture. {Shared autonomy approaches like \cite{dragan2013policy, NewmanHARMONIC2018, javdani2018shared} are not suitable within this setting because: (a) the goals lie in continuous regions and (b) the user needs to control both the \textit{goal} that the robot reaches for and the \textit{style} of the reaching trajectory (e.g., pouring, or keeping upright)}. Hence, we compare our latent action method against {a switching teleoperation strategy \textit{currently used by assistive robots}} \cite{herlant2016assistive}, where the joystick inputs alternatively control the position and orientation of the robot's end-effector.

\smallskip

\noindent\textbf{Independent Variables.} We tested two teleoperation strategies: \textit{End-Effector} and \textit{cVAE}. Under \textit{End-Effector} the user inputs applied a $6$-DoF twist to the robot's end-effector, controlling its linear and angular velocity. Participants interacted with two $2$-DoF joysticks, and were given a button to toggle between linear and angular motion \cite{herlant2016assistive, NewmanHARMONIC2018, javdani2018shared}. By contrast, in \textit{cVAE} the participants could only interact with one $2$-DoF joystick, i.e., the latent action was $z = [z_1, z_2] \in \mathbb{R}^2$. We trained the \textit{cVAE} model using state-action pairs from kinesthetic demonstrations, where we guided the robot along related sub-tasks such as reaching for the shelf, pouring objects into the bowl, and stirring. The \textit{cVAE} was trained with less than $7$ minutes of demonstration data.

\smallskip

\noindent\textbf{Dependent Measures -- Objective.} We measured the total amount of time it took for participants to complete the entire cooking task (\textit{Completion Time}), as well as the magnitude of their inputs (\textit{Joystick Input}).

\smallskip

\noindent\textbf{Dependent Measures -- Subjective.} We administered a $7$-point Likert scale survey after each condition. Questions were separated into six scales, such as ease of performing the task (\textit{Ease}) {and consistency of the controller (\textit{Consistent})}. Once users had completed the task with both strategies, we asked comparative questions about which they preferred (\textit{Prefer}), which was \textit{Easier}, and which was more \textit{Natural}.

\smallskip

\noindent\textbf{Hypotheses.} We had the following hypotheses: 
\begin{displayquote}
\textbf{H4.} \emph{Users controlling the robot arm with low-DoF latent actions will complete the cooking task more quickly and with less overall effort.}
\end{displayquote}
\begin{displayquote}
\textbf{H5.} \emph{Participants will perceive the robot as easier to work with in the cVAE condition, and will prefer the cVAE over End-Effector teleoperation.}
\end{displayquote}

\smallskip

\noindent\textbf{Experimental Setup.} We designed a cooking task where the person is making a simplified ``apple pie.'' As shown in Fig.~\ref{fig:cooking1}, the assistive robot must sequentially pour eggs, flour, and an apple into the bowl, dispose of their containers, and stir the mixture. The user sat next to the robot and controlled its behavior with a handheld joystick.

\smallskip

\noindent\textbf{Participants and Procedure.} We used a within-subjects design and counterbalanced the order of our two conditions. Eleven members of the Stanford University community ($4$ female, age range $27.4 \pm 11.8$ years) provided informed consent to participate in this study. Four subjects had prior experience interacting with the robot used in our experiment.

Before starting the study, participants were shown a video of the cooking task. Participants then separately completed the three parts of the task as visualized in Fig.~\ref{fig:cooking1}; we reset the robot to its home position between each of these sub-tasks. After the user completed these sub-tasks, we re-arranged the placement of the recycling and bowl, and users performed the entire cooking task without breaks. Participants were told about the joystick interface for each condition, and could refer to a sheet that labelled the joystick inputs.

\smallskip

\begin{figure}[t]
	\begin{center}
		\includegraphics[width=0.6\columnwidth]{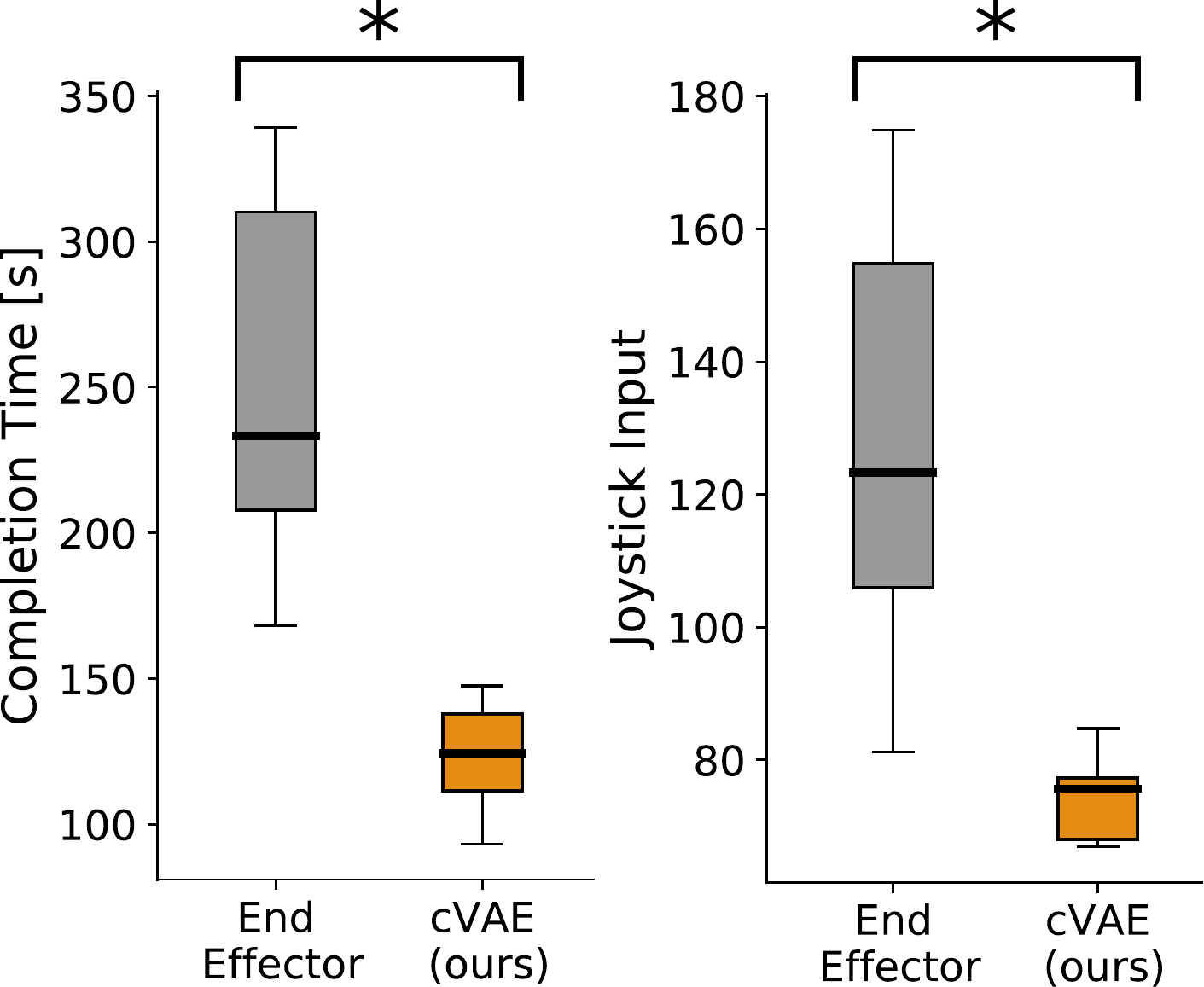}

		\caption{Objective results from the cooking user study. These results were collected on the full task (combining each of the sub-tasks from Fig.~\ref{fig:cooking1}).}

		\label{fig:cooking2}
	\end{center}
	
	\vspace{-2.5em}
	
\end{figure}

\noindent\textbf{Results -- Objective.} Our objective results are summarized in Fig.~\ref{fig:cooking2}. When using \textit{cVAE} to complete the entire recipe, participants finished the task in less time ($t(10)=-6.9$, $p<.001$), and used the joystick less frequently ($t(10)=-5.1$, $p<.001$) as compared to direct \textit{End-Effector} teleoportation.

\smallskip

\begin{figure}[t]
\vspace{0.45em}
	\begin{center}
		\includegraphics[width=0.9\columnwidth]{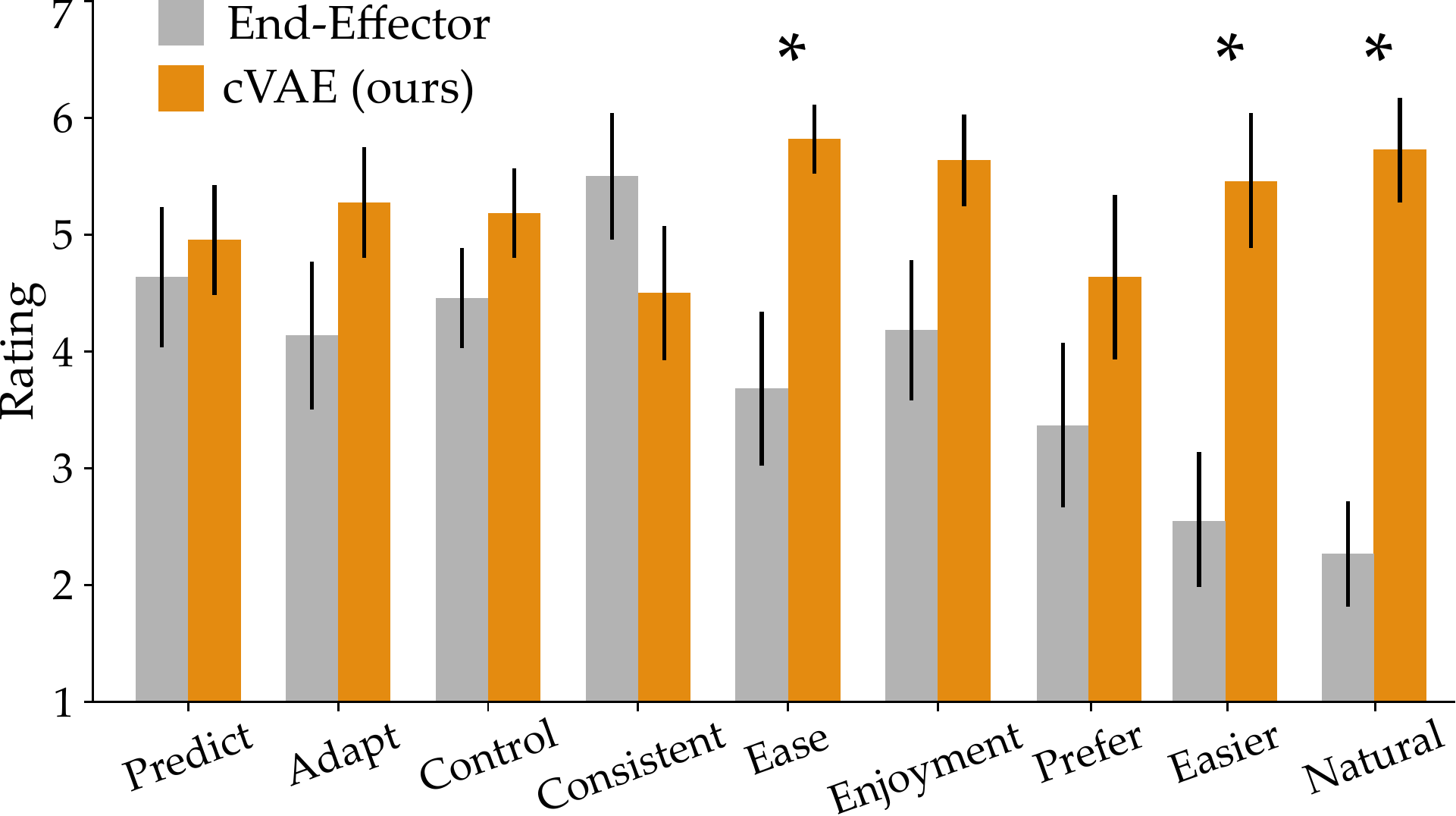}

		\caption{Subjective results from the cooking user study. {Higher ratings indicate participant agreement. Participants thought our approach required less effort (\textit{Ease}), made it easier to complete the task (\textit{Easier}), and produced more natural robot motion (\textit{Natural}) as compared to End-Effector control.}}

		\label{fig:cooking3}
	\end{center}
	
	\vspace{-2em}
	
\end{figure}

\noindent\textbf{Results -- Subjective.} We display the results of our $7$-point Likert scale surveys in Fig.~\ref{fig:cooking3}. Before reporting these results, we first confirmed the reliability of our scales. We then leveraged paired t-tests to compare user ratings for \textit{End-Effector} and \textit{cVAE} conditions. Participants perceived \textit{cVAE} as requiring less user effort ($t(10)=2.7$, $p<.05$) than \textit{End-Effector}. Participants also indicated that it was easier to complete the task with \textit{cVAE} ($t(10)=2.5$, $p<.05$), and that \textit{cVAE} caused the robot to move more naturally ($t(10)=3.8$, $p<.01$). The other scales were not significantly different.

\smallskip

\noindent\textbf{Summary.} Taken together, these results partially support our hypotheses. When controlling the robot with latent actions, users completed the cooking task more quickly and with less effort (H4). Participants believed that the \textit{cVAE} approach led to more natural robot motion, and indicated that it was easier to perform the task with latent actions. However, participants did not indicate a clear preference for either strategy (H5).

\section{Discussion and Conclusion}

\noindent\textbf{Summary.} We focused on assistive robotics settings where {the user interacts with low-Dof control interfaces}. In these settings, we showed that intelligent robots can embed their high-DoF, dextrous behavior into low-DoF latent actions for the human to control. We evaluated five different models for learning the latent actions, and determined that autoencoders conditioned on the system state accurately reconstructed the human's intended action, and also produced controllable, consistent, and scalable latent spaces.

One key advantage to latent actions is that---unlike comparable shared autonomy approaches---they can assist the human during tasks with either {discrete or continuous goal spaces}. We validated this in our two user studies. In the first (discrete), latent actions resulted in higher success than shared autonomy baselines. In the second (continuous), {participants controlled both the robot's goals \textit{and movement style}: compared against a teleoperation strategy currently employed by assistive arms}, latent actions led to improved objective and subjective performance.

\smallskip

\noindent\textbf{How practical is this approach?} In our cooking user study, the robot was trained with less than $7$ minutes of kinesthetic demonstrations. We attribute this data efficiency in part to the simplicity of our model structure: we used standard cVAEs that we trained within the robot's on-board computer. We believe this makes our approach very efficient, accurate, and easy to use in practice as compared to alternatives.

\smallskip

\noindent\textbf{Limitations and Future Work.} Although the latent actions were intuitive when the robot's state was near the training distribution, once the robot reached configurations where we had not provided demonstrations the latent actions became erratic. For example, one participant unintentionally rotated the arm upside-down when trying to pour flour, and was unable to guide the robot back towards the cooking task. We were also surprised that users did not clearly prefer the latent action approach despite its improved performance. In their questionnaire responses, participants indicated that---while it was easier to work with latent actions---they enjoyed the increased freedom of control under end-effector teleoperation.

These limitations suggest that robots should not always rely on latent actions. Future work will focus on interweaving learned latent actions with standardized teleoperation strategies, so that robots can intelligently decide when latent actions are helpful. Overall, our research aims to enable humans to seamlessly collaborate with assistive robots.

\begin{figure*}[t]
\vspace{0.25em}
	\begin{center}
		\includegraphics[width=1.8\columnwidth]{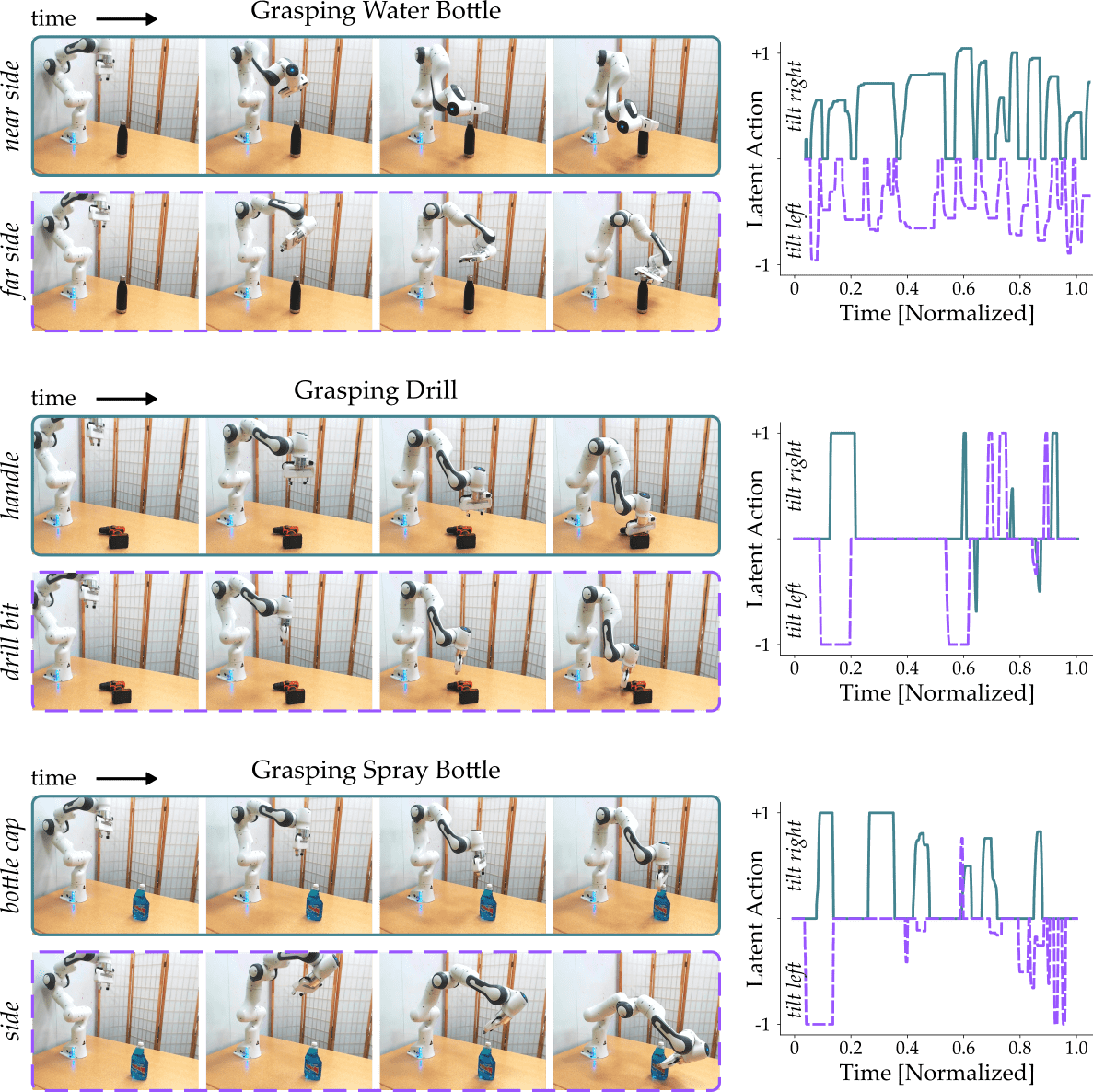}

		\caption{Example grasps with \textit{cVAE}. On the left we show the robot's trajectory, and on the right we plot the user's 1-DoF inputs over time. The robot autonomously completed the high-level reaching task, while the human's inputs controlled the robot's \textit{grasp style} along a continuous spectrum.}

		\label{fig:grasp}
	\end{center}
\vspace{-2em}
\end{figure*}


\bibliographystyle{IEEEtran}
\bibliography{IEEEabrv,BibFile}

\section{Appendix: Dexterous Manipulation}

In addition to the two user studies reported in Section~\ref{sec:userstudy}, we also applied our latent action approach to three dexterous manipulation tasks. Here the robot knows the human's high-level goal---to grasp an object---but the robot does not know the human's low-level preference for \textit{how} it should grasp that object. These dexterous manipulation tasks are an extension of the cooking task in Section~\ref{sec:cooking}, and further demonstrate that we can go beyond discrete goals to consider continuous embeddings and styles. Videos of these experiments can be found at: \href{https://youtu.be/LVzaC8w6HXU}{\color{blue}{https://youtu.be/LVzaC8w6HXU}}.

\smallskip

\noindent \textbf{Experimental Setup.} We performed these experiments using a 7-DoF Panda robotic arm (Franka Emika), and considered three different manipulation tasks: grasping a \textit{Water Bottle}, a \textit{Drill}, or a \textit{Spray Bottle}. During the offline training phase, we provided kinesthetic demonstrations that guided the robot from the start region to the target object\footnote{We provided less than $2$ minutes of kinesthetic demonstrations for each manipulation task.}. Importantly, each of the demonstrated trajectories had a slightly different \textit{grasp style}. For instance, when reaching for the drill, we provided a spectrum of kinesthetic demonstrations that grabbed the drill by the base, the handle, and the bit. We then embedded the demonstrated state-action pairs into a 1-DoF latent space using the cVAE models from our user studies.

\smallskip

\noindent \textbf{Procedure.} A single participant used a joystick to perform dexterous manipulation tasks with each object. Because we embedded to a 1-DoF latent space, the human was only able to provide inputs by pressing left or right on the joystick: our learned latent space was responsible for mapping this stream of inputs to the participant's intended grasp.

\smallskip

\noindent \textbf{Results.} In Fig.~\ref{fig:grasp} we show two example trajectories for each object when teleoperating the robot under our learned latent action approach. Similar to \cite{pervez2019motion}, we found the robot learned an embedding that \textit{arbitrates} responsibility over the task. Regardless of the user's input, the robot autonomously performed the high-level reaching motion towards the target object. Here the human's 1-DoF input controlled the robot's fine-grained grasp pose: for example, when reaching for the \textit{Drill}, tilting right (i.e., $z = +1$) guided the robot towards a grasp at the base, while tilting left (i.e., $z = -1$) guided the robot towards a grasp at the bit.

\smallskip

\noindent \textbf{Summary.} These results support our findings from the cooking task in Section~\ref{sec:cooking}, and further indicate that users can leverage our latent action approach to collaborate with assistive robots in dexterous manipulation tasks. Specifically, the robot learned to complete the high-level task autonomously, while the human dictated \textit{how} that task was performed. We emphasize that the joystick input controls the style in which the robot performs its task \textit{along a continuous spectrum}, and does not simply select between a set of discrete grasps (i.e., either grasp A or grasp B).

\end{document}